\documentclass[sigconf]{acmart}

\AtBeginDocument{%
  }

\setcopyright{acmlicensed}
\copyrightyear{2025}
\acmYear{2025}
\acmDOI{XXXXXXX.XXXXXXX}

\acmConference[SIGCSE TS '25]{The Technical Symposium on Computer Science Education (SIGCSE TS)}{Feb 26--Mar 01,
  2025}{Pittsburgh, PA}

\acmISBN{978-1-4503-XXXX-X/18/06}

\usepackage{enumitem}
\usepackage{soul}
\usepackage{pgf-pie}
\usepackage{pgfplots}

\newcommand{\ie}{\textit{i.e.},\ }
\newcommand{\eg}{\textit{e.g.},\ }
\newcommand{\etal}{\textit{et al.} }
\newcommand{\etc}{{\em etc.}}
\newcommand{\totalPapers}[0]{125}

\usepackage[most]{tcolorbox}

\tcbset{
  parskip/.style={
    before={\par\pagebreak[0]\parindent=0pt},
    after={\parfillskip=0pt\par},
  },
}
\newtcolorbox{resultbox}[1][]{%
    colback=black!3,
    colframe=black!3,
    notitle,
    sharp corners,
    borderline west={2pt}{0pt}{gray!80!black},
    enhanced,
    breakable,
    boxsep=0pt,
    left=4pt,right=2pt,top=2pt,bottom=2pt,
    }

\begin{document}

\title{Large Language Models in Computer Science Education:\\ A Systematic Literature Review}


\author{Nishat Raihan}
\email{mraihan2@gmu.edu}
\orcid{0000-0001-6242-398X}
\affiliation{%
  \institution{George Mason University}
  \city{Fairfax}
  \state{VA}
  \country{USA}
}

\author{Mohammed Latif Siddiq}
\email{msiddiq3@nd.edu}
\orcid{0000-0002-7984-3611}
\affiliation{%
  \institution{University of Notre Dame}
  \city{Notre Dame}
  \state{IN}
  \country{USA}
}

\author{Joanna C. S. Santos}
\email{joannacss@nd.edu}
\orcid{0000-0001-8743-2516}
\affiliation{%
  \institution{University of Notre Dame}
  \city{Notre Dame}
  \state{IN}
  \country{USA}
}

\author{Marcos Zampieri}
\email{mzampier@gmu.edu}
\orcid{0000-0002-2346-3847}
\affiliation{%
  \institution{George Mason University}
  \city{Fairfax}
  \state{VA}
  \country{USA}
}


\begin{abstract}
    Large language models (LLMs) are becoming increasingly better at a wide range of Natural Language Processing tasks (NLP), such as text generation and understanding. Recently, these models have extended their capabilities to coding tasks, bridging the gap between natural languages (NL) and programming languages (PL). Foundational models such as the \textit{Generative Pre-trained Transformer (GPT)} and \textit{LLaMA} series have set strong baseline performances in various NL and PL tasks. Additionally, several models have been fine-tuned specifically for code generation, showing significant improvements in code-related applications. Both foundational and fine-tuned models are increasingly used in education, helping students write, debug, and understand code. We present a comprehensive systematic literature review to examine the impact of LLMs in computer science and computer engineering education. We analyze their effectiveness in enhancing the learning experience, supporting personalized education, and aiding educators in curriculum development. We address five research questions to uncover insights into how LLMs contribute to educational outcomes, identify challenges, and suggest directions for future research. 
\end{abstract}





\maketitle

\section{Introduction}

Recent advances in Generative AI and LLMs, exemplified by GitHub Copilot \cite{denny2023conversing} and ChatGPT \cite{openai2023gpt4}, highlight their promising ability to address complex problems with human-like expertise. These advancements have a significant impact on education, where students may either benefit from or misuse these tools, compromising the integrity and quality of education \cite{phung2023generative}. This issue is particularly important in introductory Computer Science (CS) courses, which are directly affected by the capabilities of LLMs \cite{zhou2024the}. 

The ability of LLMs to efficiently handle programming tasks allows them to successfully complete assignments typically given in beginner courses, making them highly attractive to students seeking effortless solutions. Researchers have been examining the role of LLMs in CS education, focusing on how these models perform with current datasets and past assignments \cite{Hicke2023AITATA}. Identifying the use of AI tools in student work is another area of interest \cite{tanay2024exploratory}. However, current methods, including plagiarism detection software, often fail to deliver reliable performance when handling output from a recently introduced LLM \cite{meyer2023chatgpt}. 

Tools powered by LLMs offer interesting opportunities to improve CS education \cite{kumar2024impact}. When used responsibly and in the right way, they can be helpful for learning by providing students with quick feedback on coding assignments and creating different code examples to make programming concepts clearer \cite{pankiewicz2023large}. Furthermore, as Generative AI tools become more common in real-world jobs~\cite{shani2023survey}, it is important to teach students about these tools in CS classes. This ensures that students are well-prepared for careers where such tools are widely used. 

With students already adopting these tools~\cite{budhiraja2024its}, we do not yet fully understand their impact on learning. Due to the many challenges and benefits these technologies bring, understanding the impact of LLMs in CS education is paramount to improving areas such as curriculum design and assessment. While recent surveys and literature reviews have been published on different topics related to LLMs, such as their use in programming exercise generation \cite{frankford2024survey}, implications to security and privacy \cite{yao2024survey}, and software development \cite{fan2023large}, to the best of our knowledge, no comprehensive survey has been published on the impact of LLMs in CS education. A couple of surveys published on related educational topics are the one by Vierhauser et al. (2024) \cite{vierhauser2024towards} that focuses on software engineering education and the one by Cambaz et al. (2024) \cite{cambaz2024use} that focuses on programming alone. 

In this paper, we fill this important gap in the literature by presenting the first comprehensive systematic literature review (SLR) investigating the impact of LLMs on CS education. We address five carefully crafted research questions (RQs) to understand how these models influence educational outcomes, pinpoint challenges, and suggest future research directions. Following Kitchenham and Charters (2007) guidelines \cite{kitchenham2007guidelines} for conducting SLR, we use a thorough search strategy across multiple databases, applying strict inclusion and exclusion criteria to ensure the studies selected are relevant and of high quality. Our results reveal the transformative impact of LLMs in CS educational practices. To enable reproducibility, our scripts, and data are available in a GitHub repository\footnote{\url{https://github.com/s2e-lab/llm-education-survey}}.

\section{Methodology}

To conduct this SLR, we follow the aforementioned guidelines \cite{kitchenham2007guidelines} that suggest three main steps: \textit{planning} the literature review, \textit{conducting} the literature review, and \textit{reporting} the results. During the planning phase, we set five clear RQs and created a detailed plan for our SLR. In the conducting phase, we search for relevant studies and select them based on specific criteria. Finally, in the reporting phase, we organize and present our findings in this paper.

\subsection{Research Questions}
\label{sec:rq}

The five RQs addressed in this SLR to understand the use and impact of LLMs in CS education are the following:

\begin{itemize}[leftmargin=24pt, topsep=0pt,itemsep=0pt]
\item[\textbf{RQ1:}]  \textit{\textbf{What are the educational levels in which LLMs are used?}}
\end{itemize}

In this RQ, we examine the education stages (\eg undergraduate level, graduate level, \etc), in which LLMs are integrated. This examination aims to identify the most effective stages for introducing and integrating these models, enhancing learning outcomes, and guiding future educational practices.

\begin{itemize}[leftmargin=24pt, topsep=0pt,itemsep=0pt]
\item[\textbf{RQ2:}]   \textit{\textbf{What are the sub-disciplines of CS that are the focus of the studied papers?}}
\end{itemize} 

We investigate what are the specific sub-disciplines (\eg CS1, software testing, \etc) that were the target of the studied works.\footnote{As described in Section~\ref{subsec:SearchMethod}, we consider some sub-disciplines that are commonly within the scope of computer engineering along with CS sub-disciplines. The acronym used throughout the paper is CS.} This RQ aims to identify areas needing further research.

\begin{itemize}[leftmargin=24pt, topsep=0pt,itemsep=0pt]
\item[\textbf{RQ3:}]  \textit{\textbf{What research methodologies are mostly used in the papers?}}
\end{itemize}

We explore the research methodologies, such as experimental designs and data analysis techniques, employed in the selected papers. This RQ aims to understand the field's research practices and standards with respect to this ML-based technology.

\begin{itemize}[leftmargin=24pt, topsep=0pt,itemsep=0pt]
\item[\textbf{RQ4:}] \textit{\textbf{What are the most commonly used programming languages (PLs) in studies involving LLMs?}}
\end{itemize}

In this RQ, we examine the programming languages that are the focus of the paper.
Identifying the frequently targeted languages helps reveal trends and gaps in educational research and practice.

\begin{itemize}[leftmargin=24pt, topsep=0pt,itemsep=0pt]
\item[\textbf{RQ5:}]  \textit{\textbf{Which large language models (LLMs) are employed in these studies?}}
\end{itemize}
In this RQ, we study the LLMs most widely used in research papers. Cataloging the specific LLMs used provides insights into the diversity and rationale for model selection.

\subsection{Search Method}\label{subsec:SearchMethod}
To answer our RQs, we use the search query below to retrieve all \textit{\textbf{primary}} works related to LLMs for CS education:

\begin{center}
\texttt{("software engineering" \textbf{OR} "programming" \textbf{OR} "software development" \textbf{OR} "computer science" \textbf{OR} "computer engineering")  \textbf{AND}  ("education" \textbf{OR} "teaching")  \textbf{AND}  ("LLM" \textbf{OR} "large language model")}
\end{center}

We apply this query to the following library databases~\footnote{\textit{portal.acm.org}. IEEE Xplore: \textit{ieeexplore.ieee.org}. Scopus: \textit{scopus.com}. ACL: \textit{aclanthology.org}. Web of Science: \textit{isiknowledge.com}. Springer: \textit{link.springer.com}. Science@Direct: \textit{sciencedirect.com}. ArXiv: \textit{arxiv.org}.}: 
the \textsf{ACM Digital Library}, 
\textsf{IEEE Xplore}, 
\textsf{Scopus}, 
the \textsf{ACL Anthology}, 
\textsf{ISI Web of Science}, 
\textsf{Springer Link},
\textsf{Science@Direct},   and \textsf{ArXiv}.
This search query results in a total of {1,735} papers.

\subsection{Inclusion and Exclusion Criteria}

We vet the papers to exclude those that do not meet our \textit{inclusion} criteria or that meet our \textit{exclusion} criteria. Our criteria,  shown in Table~\ref{table:criteria}, ensure the relevance and quality of the selected works. 

\begin{table}[!ht]
\small
\caption{Inclusion and Exclusion criteria to Select Papers}\label{table:criteria}
\begin{tabular}{@{}ll@{}}
\toprule
\multicolumn{1}{c}{\textbf{Inclusion Criteria}}  & \multicolumn{1}{c}{\textbf{Exclusion Criteria}}  \\ \midrule
\begin{tabular}[c]{@{}p{4cm}@{}}
    \textbf{I1} Full papers (\ie at least 4 full pages of text, excluding references). \\
    \textbf{I2} Written in English. \\
    \textbf{I3} Focus on or investigate the use of code LLMs to teach computing concepts. \\
    \textbf{I4} Written between January 2019 - June 2024.   
\end{tabular} & 
\begin{tabular}[c]{@{}p{4cm}@{}}
    \textbf{E1} Duplicated studies\\  
    \textbf{E2} Not written in English. \\
    \textbf{E3} Abstracts, posters, or extended abstracts with less than 4 full pages of text. \\ 
    \textbf{E4}: Survey and Systematic Literature Reviews (SLRs).
\end{tabular} \\ \bottomrule
\end{tabular}
\end{table}

We begin with a total of {1,735} primary studies. 
First, we removed duplicated studies and papers not published within the past 5 years, obtaining a total of {1,423} papers.
Subsequently, we inspect each paper's \textit{title}, \textit{keywords}, \textit{number of pages}, and \textit{abstract} to determine their relevance based on our inclusion and exclusion criteria. This screening process reduces the number of papers to {187}. Finally, we apply the same criteria to the full text of these papers, leaving us with \textbf{\totalPapers} papers included in this literature review.

\subsection{Data Extraction}

As we reviewed the papers, we extracted the key information we were looking for to answer our RQs: 
the \textit{educational level},  the \textit{CS discipline}, and \textit{programming languages} that were the focus of the paper as well as the \textit{LLMs} and \textit{research methodologies} that were employed.
This data was extracted by two of the authors and peer-reviewed by the senior author.

\section{Results}

Upon carefully reviewing the \totalPapers~selected papers, we conducted a high-level analysis addressing the RQs presented in Section \ref{sec:rq}. 

\subsection{RQ1: Educational Levels} \label{subsec:rq1}

\begin{figure}[!htbp]
    \centering
    \includegraphics[width=1\linewidth]{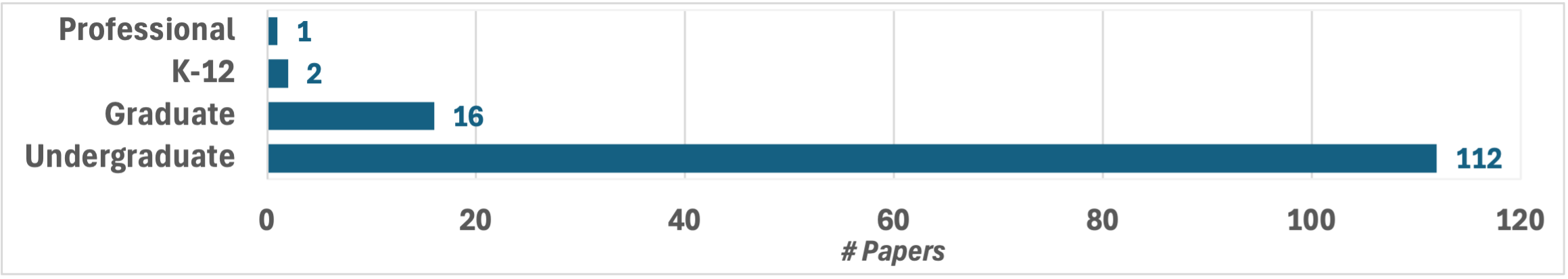}
    \caption{Educational Levels where LLMs are used.}
    \label{fig:edu_level}
\end{figure}

As shown in Figure~\ref{fig:edu_level}, \textbf{111}  of the studied papers focus on \textit{{undergraduate-level}} CS courses~\cite{abolnejadian2024leveraging,agarwal2024which,anishka2023can,arora2024analyzing,aviv2024impact,azaiz2023aienhanced,azaiz2024feedbackgeneration,bakas2024integrating,balse2023evaluating,balse2023investigating,becker2023programming,berrezuetaguzman2023recommendations,budhiraja2024its,bukar2024text,cao2023scaffolding,cipriano2023gpt3,cipriano2024llms,delcarpiogutierrez2024evaluating,dengel2023qualitative,denny2023conversing,denny2024desirable,denny2024prompt,doughty2024comparative,drori2023human,ellis2024chatgpt,fan2023exploring,farah2023prompting,feng2024more,fernandez2024cs1,frankford2024aitutoring,freire2024may,garg2024impact,grevisse2024comparative,grevisse2024docimological,gumina2023teaching,haindl2024students,hanifi2023chatgpt,hoq2024detecting,hou2024effects,jacobs2024evaluating,jin2024teach,jordan2024need,joshi2024chatgpt,jost2024impact,jury2024evaluating,karnalim2024detecting,kazemitabaar2023studying,kazemitabaar2024codeaid,kiesler2023large,kim2024chatgpt,kirova2024software,kosar2024computer,koutcheme2023training,kruger2024performance,kumar2024using,kuramitsu2023kogi,lau2023ban,leinonen2023comparing,li2023evaluating,li2023potential,liffiton2024codehelp,liu2024beyond,liu2024teaching,lyu2024evaluating,ma2024teach,macneil2023experiences,manley2024examining,mendoncca2024evaluating,nguyen2024beginning,oli2024automated,oosterwyk2024beyond,orenstrakh2023detecting,padiyath2024insights,pankiewicz2023large,parker2024large,piccolo2023evaluating,prakash2024integrating,prather2023robots,prather2024interactions,qureshi2023chatgpt,raihan2024cseprompts,rajala2023call,rasnayaka2024empirical,reiche2024bridging,rodriguezecheverria2024analysis,sanchez2023assessing,sarsa2022automatic,sarshartehrani2024enhancing,savelka2023efficient,savelka2023thrilled,savelka2024gpt3,scholl2024analyzing,sharpe2024can,sheese2024patterns,song2024automatic,sterbini2024automated,strzelecki2024acceptance,ta2023exgen,tanay2024exploratory,tran2023generating,tu2023should,vadaparty2024cs1llm,venkatesh2023evaluating,wan2024automated,wang2023exploring,wang2024enhancing,wolfer2024qualitative,xiao2024qacp,zastudil2023generative,zhang2023students,zhang2024assistant}. While \textbf{15} works explored advanced courses typically taught at the {\textit{graduate level}}~\cite{arora2024analyzing,becker2023programming,bien2024generative,bukar2024text,drori2023human,estevezayres2024evaluation,gehringer2024dualsubmission,hanifi2023chatgpt,jalil2023chatgpt,parker2024large,prather2023robots,shen2024implications,strzelecki2024acceptance,tu2023should,zheng2023chatgpt}, only  \textbf{4} papers include {\textit{PhD-level}} courses~\cite{arora2024analyzing,bukar2024text,gehringer2024dualsubmission,prather2023robots}, and just \textbf{2} papers addressed {\textit{K-12}} education \cite{grover2024teaching, kazemitabaar2024novices}. There was one work~\cite{petrovska2024incorporating} that examined how ChatGPT is used as a means for training employees in a {\textit{software engineering workplace}} (professional context).

This strong focus on undergraduate level is primarily due to the limited presence of CS courses at lower educational levels (K-12) and the current limitations of most LLMs in handling higher-level content \cite{wolfer2024qualitative, jalil2023chatgpt}. Although newer models like GPT-4 have shown promising results for some graduate courses \cite{piccolo2023evaluating, drori2023human}, many earlier studies \cite{sanchez2023assessing, jalil2023chatgpt} did not have the chance to test these capabilities. Even in more recent works, GPT-4 is sometimes excluded due to its cost or other practical issues \cite{rodriguezecheverria2024analysis}.

\begin{resultbox}
\textbf{RQ1 Results:} 
Over \textbf{80\%} of the papers focused on undergraduate-level CS education. Further studies 
are needed to assess the effectiveness and helpfulness of LLMs for graduate students and professionals undergoing CS training.
\end{resultbox}

\subsection{RQ2: CS Sub-disciplines}


As shown in Table \ref{tab:disciplines}, over half of the \totalPapers{} observed studies focus on \textsl{{introduction to programming}}, predominantly in Python and occasionally in Java. This emphasis is expected, given that CS students frequently use LLMs for code generation \cite{bukar2024text}. The emphasis on introductory programming arises from Python's widespread adoption as the first language in many CS curricula and Java's significant role in teaching object-oriented programming principles.

\begin{table}[!htbp]
    \small
    \centering
    \caption{CS Disciplines Explored by the Studied Papers.}
    \begin{tabular}{@{}l@{}c@{ }p{4cm}@{}}
        \toprule
        \textbf{CS Discipline} & \textbf{Total} & \textbf{References} \\
        \midrule
		Introduction to Programming & 65 & \cite{abolnejadian2024leveraging,agarwal2024which,ahmed2024potentiality,anishka2023can,aviv2024impact,azaiz2023aienhanced,azaiz2024feedbackgeneration,bakas2024integrating,balse2023evaluating,balse2023investigating,berrezuetaguzman2023recommendations,budhiraja2024its,cao2023scaffolding,delcarpiogutierrez2024evaluating,denny2023conversing,denny2024desirable,denny2024prompt,doughty2024comparative,ellis2024chatgpt,fan2023exploring,fernandez2024cs1,frankford2024aitutoring,haindl2024students,hoq2024detecting,jacobs2024evaluating,jordan2024need,joshi2024chatgpt,jost2024impact,jury2024evaluating,karnalim2024detecting,kazemitabaar2023studying,kazemitabaar2024codeaid,kazemitabaar2024novices,kiesler2023large,kosar2024computer,koutcheme2023training,kumar2024using,lau2023ban,leinonen2023comparing,liu2024beyond,liu2024teaching,lyu2024evaluating,ma2024teach,manley2024examining,nguyen2024beginning,oli2024automated,oosterwyk2024beyond,padiyath2024insights,prather2024interactions,raihan2024cseprompts,roest2024nextstep,sanchez2023assessing,sarsa2022automatic,sarshartehrani2024enhancing,savelka2023efficient,savelka2023thrilled,savelka2024gpt3,scholl2024analyzing,sharpe2024can,sheese2024patterns,ta2023exgen,venkatesh2023evaluating,wan2024automated,xiao2024qacp,zhang2023students} \\
		Introduction to CS & 19 & \cite{budhiraja2024its,denny2023conversing,denny2024prompt,doughty2024comparative,grevisse2024comparative,grevisse2024docimological,joshi2024chatgpt,manley2024examining,oosterwyk2024beyond,orenstrakh2023detecting,sarshartehrani2024enhancing,savelka2023thrilled,savelka2024gpt3,song2024automatic,tran2023generating,vadaparty2024cs1llm,wang2023exploring,wang2024enhancing,wolfer2024qualitative} \\
		Data Science & 9 & \cite{bien2024generative,garg2024impact,kruger2024performance,kuramitsu2023kogi,liffiton2024codehelp,shen2024implications,tu2023should,wolfer2024qualitative,zheng2023chatgpt} \\
		Software Engineering & 8 & \cite{farah2023prompting,hanifi2023chatgpt,kirova2024software,kruger2024performance,petrovska2024incorporating,rasnayaka2024empirical,rodriguezecheverria2024analysis,tanay2024exploratory} \\
		Object-Oriented Programming & 4 & \cite{cipriano2023gpt3,cipriano2024llms,kruger2024performance,pankiewicz2023large} \\
		Algorithms & 4 & \cite{jin2024teach,kuramitsu2023kogi,mendoncca2024evaluating,sterbini2024automated} \\
		Web Development & 3 & \cite{kruger2024performance,macneil2023experiences,rajala2023call} \\
		Machine Learning & 3 & \cite{drori2023human,reiche2024bridging,wolfer2024qualitative} \\
		Computer architecture & 3 & \cite{gehringer2024dualsubmission,mendoncca2024evaluating,zhang2024assistant} \\
        \midrule
        \textit{CS Education in General} & 10 & \cite{becker2023programming,bukar2024text,dengel2023qualitative,grover2024teaching,li2023potential,parker2024large,prather2023robots,qureshi2023chatgpt,strzelecki2024acceptance,zastudil2023generative} \\		
        \midrule
        \textit{Others} & 14 & \cite{arora2024analyzing,kruger2024performance, agarwal2024which,kruger2024performance, kruger2024performance, estevezayres2024evaluation, jalil2023chatgpt, gumina2023teaching, feng2024more, freire2024may, prakash2024integrating, mendoncca2024evaluating, piccolo2023evaluating, li2023evaluating, kim2024chatgpt}\\ 
	\bottomrule
    \end{tabular}
    \label{tab:disciplines}
\end{table}

There were \textbf{19} works that focused on \textit{{introductory CS}} concepts, featuring Q\&As and multiple-choice questions related to basic computer science topics \cite{zhang2024assistant, wang2024enhancing, grevisse2024comparative}. These studies not only assess LLM-generated code, solutions, and feedback but also explore the generation of tasks, assignments, and questions \cite{pankiewicz2023large, cipriano2023gpt}. More advanced courses, such as \textit{Data Science}, present mixed findings regarding the effectiveness of LLMs; some studies claim that LLMs perform well \cite{bien2024generative}, while others disagree \cite{zheng2023chatgpt, tu2023should}. 

Due to space constraints, disciplines with less than 3 papers are aggregated as ``Others'' in Table~\ref{tab:disciplines}. 
These other advanced topics were
\textit{{Distributed Systems}}~\cite{arora2024analyzing,kruger2024performance},
\textit{{Operating Systems}}~\cite{agarwal2024which,kruger2024performance},
\textit{{Computer Networks}}~\cite{kruger2024performance},
\textit{{Numerical Analysis}}~\cite{kruger2024performance},
\textit{{Interactive Systems}} ~\cite{kruger2024performance},
\textit{{Real-Time Systems}} ~\cite{kruger2024performance},
\textit{{Concurrent, Parallel and Distributed Computing}} ~\cite{estevezayres2024evaluation},
\textit{{Software Testing}} ~\cite{jalil2023chatgpt},
\textit{{Information Technology}} ~\cite{gumina2023teaching},
\textit{{Computer Graphics}} ~\cite{feng2024more},
\textit{{Human-computer Interaction}} ~\cite{freire2024may},
\textit{{Databases}} ~\cite{prakash2024integrating},
\textit{{Automata Theory and Formal Languages}} ~\cite{mendoncca2024evaluating},
\textit{{Bioinformatics}} ~\cite{piccolo2023evaluating},
\textit{{Software Security}} ~\cite{li2023evaluating}, and
\textit{{Data Visualization}} ~\cite{kim2024chatgpt}.

\begin{resultbox}
\textbf{RQ2 Results:} \textbf{67\%} of works focus on \textit{introduction to programming} and \textit{introduction to CS}. There is limited focus on more advanced CS concepts, suggesting a need for further exploration on LLMs' ability in helping to teach advanced CS concepts.
\end{resultbox}

\subsection{RQ3: Research Methodologies}


Table \ref{tab:methodologies} summarizes the research methodologies followed by the studied papers.
Our findings show that \textbf{38\%} of papers use case studies and ethnography as their research method, which means these papers are focused on how LLMs can be used in different use cases of CS education. Moreover, 24\% of the papers used the action research method, where the researchers introduced novel LLM-based tools and techniques and applied them in a CS educational context. We also found \textbf{24} works in which researchers did experiments with students from various levels as described in RQ1 (Section \ref{subsec:rq1}). In 14\% of the works we analyzed, we found researchers were involved in data-centric analysis, and in 12\% of cases, they used grounded theory for qualitative analysis. 12\% of papers were involved in engineering research that invents and evaluates LLM-based artifacts for CS education. Researchers were also involved in interview studies and case studies. They also used multi-methodology and mixed-methods research to use LLMs in CS education. In 5\% of works, they benchmarked LLMs for CS education tasks. The rest of the two works used  Optimization Studies \cite{pankiewicz2023large} and Repository Mining \cite{xiao2024qacp} as their research methodology. 

\begin{table}[!htbp]
    \small
    \centering
    \caption{Research Methodologies used by the Papers.}
    \begin{tabular}{@{}p{3cm}@{}cp{4.5cm}@{}}
        \toprule
        \textbf{Methodology} & \textbf{Total} & \textbf{References} \\
        \midrule
		Case Study and Ethnography & 48 & \cite{abolnejadian2024leveraging,ahmed2024potentiality,bakas2024integrating,balse2023evaluating,balse2023investigating,berrezuetaguzman2023recommendations,bien2024generative,cao2023scaffolding,cipriano2024llms,denny2024prompt,drori2023human,fan2023exploring,farah2023prompting,feng2024more,frankford2024aitutoring,freire2024may,grevisse2024comparative,grevisse2024docimological,grover2024teaching,haindl2024students,hou2024effects,jin2024teach,jordan2024need,jury2024evaluating,kazemitabaar2023studying,kazemitabaar2024codeaid,kiesler2023large,kruger2024performance,kumar2024using,li2023evaluating,liffiton2024codehelp,liu2024beyond,liu2024teaching,mendoncca2024evaluating,mezzaro2024empirical,pankiewicz2023large,piccolo2023evaluating,prakash2024integrating,roest2024nextstep,sanchez2023assessing,savelka2023efficient,shen2024implications,ta2023exgen,tu2023should,vadaparty2024cs1llm,venkatesh2023evaluating,wolfer2024qualitative,xiao2024qacp} \\
		Action Research & 30 & \cite{anishka2023can,balse2023investigating,berrezuetaguzman2023recommendations,cipriano2023gpt3,cowan2024enhancing,delcarpiogutierrez2024evaluating,doughty2024comparative,fan2023exploring,farah2023prompting,fernandez2024cs1,frankford2024aitutoring,gumina2023teaching,hoq2024detecting,jin2024teach,joshi2024chatgpt,karnalim2024detecting,kazemitabaar2023studying,kazemitabaar2024codeaid,kosar2024computer,li2023evaluating,ma2024teach,macneil2023experiences,petrovska2024incorporating,roest2024nextstep,sarshartehrani2024enhancing,scholl2024analyzing,song2024automatic,wang2024enhancing,zhang2024assistant,zheng2023chatgpt} \\
		Experiments with Human Participants & 24 & \cite{abolnejadian2024leveraging,ahmed2024potentiality,azaiz2023aienhanced,azaiz2024feedbackgeneration,balse2023evaluating,dengel2023qualitative,doughty2024comparative,garg2024impact,grover2024teaching,haindl2024students,kazemitabaar2024codeaid,kazemitabaar2024novices,koutcheme2023training,kruger2024performance,lau2023ban,li2023evaluating,li2023potential,oli2024automated,savelka2023efficient,scholl2024analyzing,strzelecki2024acceptance,vadaparty2024cs1llm,wang2023exploring,zhang2023students} \\
		Data Science & 18 & \cite{azaiz2024feedbackgeneration,bien2024generative,budhiraja2024its,bukar2024text,cipriano2024llms,denny2023conversing,doughty2024comparative,estevezayres2024evaluation,garg2024impact,grevisse2024docimological,jacobs2024evaluating,koutcheme2023training,kumar2024using,oli2024automated,parker2024large,scholl2024analyzing,sterbini2024automated,strzelecki2024acceptance} \\
		Grounded Theory & 15 & \cite{agarwal2024which,azaiz2023aienhanced,becker2023programming,denny2024desirable,denny2024prompt,jost2024impact,kirova2024software,kosar2024computer,lau2023ban,manley2024examining,mezzaro2024empirical,nguyen2024beginning,reiche2024bridging,sheese2024patterns,tanay2024exploratory} \\
		Engineering Research (Design Science) & 15 & \cite{bakas2024integrating,cao2023scaffolding,denny2023conversing,ellis2024chatgpt,jalil2023chatgpt,kuramitsu2023kogi,leinonen2023comparing,liffiton2024codehelp,prather2024interactions,rajala2023call,rodriguezecheverria2024analysis,sarsa2022automatic,savelka2023thrilled,tran2023generating,wan2024automated} \\
		Qualitative Surveys & 9 & \cite{arora2024analyzing,hanifi2023chatgpt,jacobs2024evaluating,joshi2024chatgpt,kim2024chatgpt,padiyath2024insights,wang2023exploring,zastudil2023generative,zhang2023students} \\
		Longitudinal Studies & 8 & \cite{budhiraja2024its,jost2024impact,lyu2024evaluating,manley2024examining,oosterwyk2024beyond,reiche2024bridging,savelka2024gpt3,tanay2024exploratory} \\
		Mixed Methods Research & 7 & \cite{aviv2024impact,feng2024more,kiesler2023large,manley2024examining,padiyath2024insights,prakash2024integrating,rasnayaka2024empirical} \\
		Benchmarking & 6 & \cite{babe2023studenteval,jalil2023chatgpt,raihan2024cseprompts,rodriguezecheverria2024analysis,savelka2023thrilled,sharpe2024can} \\
		Case Survey & 4 & \cite{gehringer2024dualsubmission,orenstrakh2023detecting,prather2023robots,qureshi2023chatgpt} \\
		Optimization Studies & 1 & \cite{pankiewicz2023large} \\
		Repository Mining & 1 & \cite{xiao2024qacp} \\
	\bottomrule
    \end{tabular}
    \label{tab:methodologies}
\end{table}

While most studies conducted interviews with students, teachers as well as practitioners (humans), the work by Dengel~\etal~\cite{dengel2023qualitative} conducted semi-structured interviews with LLMs. Their goal was to examine the applicability of qualitative research methods to interviews with LLMs. In their study, LLMs were asked questions related to the relevance of computer science in K-12 education.

\begin{resultbox}
\textbf{RQ3 Results:} Around \textbf{60\%} works focus on case studies of using LLMs and action research for creating tools around LLMs for CS educational tasks. Researchers also did qualitative research by conducting surveys. Limited works explore data-centric analysis and benchmarking LLMs.
\end{resultbox}

\subsection{RQ4: Programming Languages}

\begin{figure}[!htbp]
    \centering
    \includegraphics[width=\linewidth]{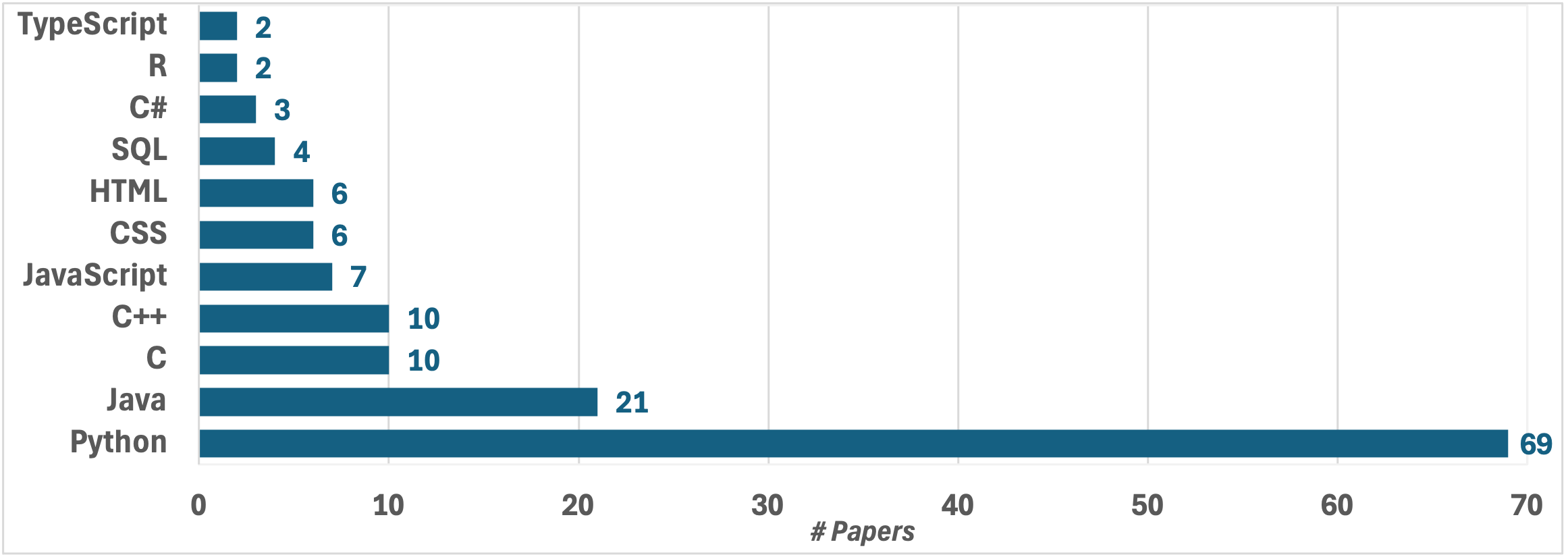}
    \caption{Top 10 Programming Languages.}
    \label{fig:languages}
\end{figure}

Figure~\ref{fig:languages} shows the top 10 programming languages that the reviewed papers focused on (\textit{TypeScript} and \textit{R} are tied for the 10th position). Among the programming languages, \textit{{Python}} is the most studied, being mentioned in \textbf{55\%} of the papers. This prominence is likely because many studies focus on introductory CS courses, where Python is often the first language taught. The second most used language is {\textit{Java}} \cite{wolfer2024qualitative, jalil2023chatgpt}, primarily taught as an object-oriented programming language. \textit{C} \cite{kazemitabaar2024codeaid} and \textit{C++} \cite{leinonen2023comparing} receive comparatively less attention. Fewer works focus on web languages like \textit{JavaScript}, \textit{HTML}, and \textit{CSS}~\cite{gumina2023teaching,jost2024impact,kim2024chatgpt,kruger2024performance,macneil2023experiences,prather2023robots,rajala2023call,prather2023robots,wolfer2024qualitative}.

Our findings also reveal that LLMs are employed not only for code generation but also for code-related Q\&A tasks \cite{zhang2024assistant, wang2024enhancing, manley2024examining, venkatesh2023evaluating} and MCQs \cite{grevisse2024comparative, savelka2024gpt3, song2024automatic, grevisse2024docimological} across various CS courses. Moreover, some studies do not specify particular tasks but instead provide a broader overview, such as examining how LLMs are perceived by students \cite{qureshi2023chatgpt} and teachers \cite{strzelecki2024acceptance, zastudil2023generative}, evaluating course feedback \cite{estevezayres2024evaluation, parker2024large, li2023potential}, and generating scaffolds \cite{cao2023scaffolding}.

\begin{resultbox}
\textbf{RQ4 Results:} 
Most papers focused on Python and Java code generation while neglecting other commonly used languages such as JavaScript, C \& C++.
\end{resultbox}

\subsection{RQ5: Most commonly used LLMs}

\begin{figure}[!htbp]
    \centering
    \includegraphics[width=\linewidth]{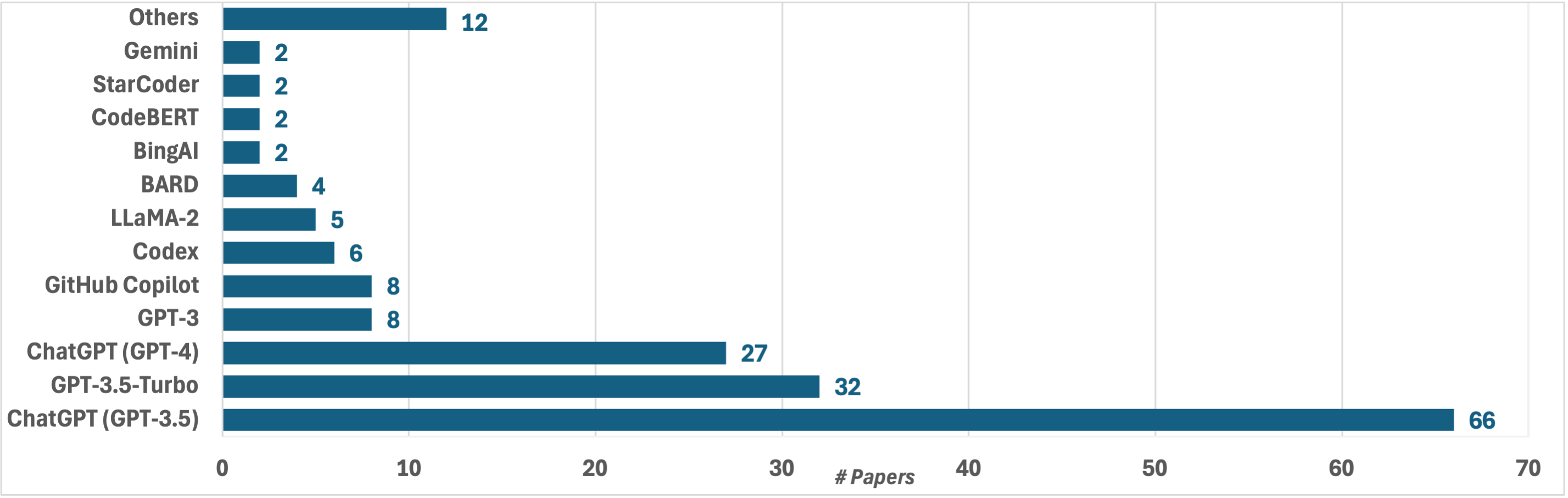}
    \caption{Most commonly used LLMs. \textit{Others} include models that are used only once. A work may use more than one LLM.}
    \label{fig:models}
\end{figure}

Most works used ChatGPT \cite{brown2020language} without employing the API, primarily relying on the earlier GPT-3.5 model, with some using GitHub Copilot (OpenAI's Codex) \cite{chen2021evaluating} and GPT-4 \cite{achiam2023gpt} (see Figure \ref{fig:models}). This trend is largely because many of the studies were conducted before the release of GPT-4 and the higher cost of GPT-4 compared to the mostly free GPT-3.5. Additionally, several works accessed GPT-3.5-Turbo through OpenAI's API. Besides OpenAI models, Microsoft's BingAI\footnote{bing.com/chat}, as well as Google's BARD\footnote{bard.google.com} and Gemini\footnote{gemini.google.com}, have been used occasionally. Code-finetuned models like StarCoder \cite{li2023starcoder} and CodeBERT \cite{feng2020codebert} were also used a few times. Unlike OpenAI's models, StarCoder and CodeBERT are free to use.

There were 26 different models used only once across 12 papers. These include Anthropic's Claude\footnote{claude.ai} and earlier LLMs like Mistral \cite{jiang2023mistral}, Falcon \cite{penedo2023refinedweb}, MPT \cite{mosaicml_mpt30b} etc. 

\begin{resultbox}
\textbf{RQ5 Results:} Most papers used commercial models in their studies, with ChatGPT  being the most used model due to its popularity among students. However, papers mostly used its older version (GPT-3.5) instead of GPT-4 due to its costs.
\end{resultbox}
\section{Discussion}

Along with the five RQs answered in this SLR, in our comprehensive analysis, we also identified four important discussion points on LLMs in CS Education as follows: 


\noindent\textbf{-- Students' and instructors' sentiment about using LLMs in CS Education:} 
Students generally have positive experiences with LLMs and LLM-based tools \cite{petrovska2024incorporating,roest2024nextstep,ahmed2024potentiality,prather2024interactions,arora2024analyzing}. Studies have shown that students consider examples generated by LLMs helpful \cite{jury2024evaluating} and perceive that LLMs could enhance their knowledge \cite{tanay2024exploratory} by providing helpful feedback \cite{zhang2023students}. CS Students praised LLMs for providing explanations that were easy to understand \cite{leinonen2023comparing} and thought that LLMs could be an additional agent with teaching assistants. However, studies have also shown that students expressed some frustration about crafting prompts that elicit the desired output \cite{nguyen2024beginning}. Furthermore, some studies indicate that students found it hard to find relevant or accurate responses from LLMs \cite{arora2024analyzing}. From the perspective of the instructors, studies have indicated that CS instructors found a negative correlation between the usage of LLMs and students' grades \cite{jost2024impact}. They found LLMs could negatively affect students’ ability to solve programming tasks independently \cite{jost2024impact}. They expressed concerns about proper learning, over-reliance on tools, and plagiarism when using LLMs in CS education \cite{kazemitabaar2024novices}.

\noindent\textbf{-- Task completion with LLMs and LLM-based tools:} 
As described in the results of RQ2 and RQ4, LLMs and LLM-based tools were applied to solve assignments and programming problems in different PLs from different courses in CS. Regarding the successful completion of the tasks, we found mixed results regarding the use of LLMs. Studies have shown that LLMs can help students solve introductory programming problems, repair buggy code \cite{koutcheme2023training,ma2024teach,wan2024automated}, and help write better code \cite{kazemitabaar2023studying,kiesler2023large,prather2024interactions}. According to the findings, LLMs are generally better at writing code than solving question-answer \cite{venkatesh2023evaluating}. They can also generate programming problems \cite{sarsa2022automatic,ta2023exgen}, MCQs \cite{song2024automatic,tran2023generating}, and detect AI-generated code despite having false positives \cite{karnalim2024detecting}. LLMs can also provide feedback to the student to improve their code \cite{parker2024large,li2023potential}. However, LLMs can partially help with data science \cite{zheng2023chatgpt,tu2023should} but hardly solve machine learning problems \cite{drori2023human}. They also suffer from problems in other languages other than English. For example, LLM performed poorly on Chinese Python question-answering problems \cite{xiao2024qacp}.

\noindent\textbf{-- Adoption of LLMs in different use cases:} 
LLMs are heavily adopted by students \cite{hanifi2023chatgpt,manley2024examining}; they often try ChatGPT to solve their problems but remain skeptical overall \cite{rajala2023call,manley2024examining}. The adoption of LLMs varied depending on the students’ coding skills and prior experience \cite{rasnayaka2024empirical}. It is also significantly influenced by their perception of future career norms \cite{padiyath2024insights}. There is potential for rapid iteration, creative ideation, and avoiding social pressures for using LLMs \cite{hou2024effects}. Students used it as a chatbot \cite{kazemitabaar2024codeaid}, integrated with the IDE \cite{kuramitsu2023kogi}, and as a substitute for the teaching assistant \cite{tanay2024exploratory}. They usually read the generated code to solve a task and mostly understand them \cite{vadaparty2024cs1llm}. 

\noindent\textbf{-- Expectation and future direction of using LLMs in CS Education:} 
Though LLMs can help solve assignments, provide feedback, and repair code, both students and instructors desire more than answers from LLMs \cite{denny2024desirable}. Students and instructors agree that LLMs should be welcome in academia \cite{kirova2024software, grover2024teaching} and that the integration of LLMs with teaching can lead to a better understanding \cite{arora2024analyzing}. Instructors ask to change the curriculum as they can solve most of the data structure problems \cite{shen2024implications} but are urged to handle LLMs carefully \cite{becker2023programming}.





\section{Conclusion}

In this paper, we presented the first comprehensive SLR on LLMs in CS education. We identified and analyzed 1,735 related papers. After applying well-defined inclusion and exclusion criteria, we described \totalPapers{} relevant papers in this SLR - the most related SLR to ours \cite{cambaz2024use} has covered 21 papers focusing only on programming. Taking the \totalPapers{} relevant papers into consideration, we answered five important RQs related to educational levels, sub-disciplines, methodologies, and PLs. We also presented a brief discussion on the adoption of LLMs, students' sentiments, and future directions. 

Our findings indicate that most current research focuses on undergraduate education and introductory programming courses. They also indicate that most research applies case-based studies, while the most widely-used PL is Python. All in all, although students are usually positive about using LLMs, instructors are worried about learning effectiveness because of potential over-reliance on them. Our SLR also indicates that educators are gradually adopting LLMs in their courses but that most CS curricula still need to be changed to accommodate recent advances in AI. We hope that the insights gained from this comprehensive SLR will help inform and enhance future research and applications of LLMs in CS education, contributing to a deeper understanding of their role and effectiveness in various educational contexts.


\bibliographystyle{unsrtnat}
\bibliography{references}



\end{document}